\documentclass[10pt,letterpaper]{article}

\usepackage{cogsci}
\usepackage{tipa}
\cogscifinalcopy 
\usepackage[hidelinks]{hyperref}
\usepackage{pslatex}
\usepackage{apacite}
\usepackage{float} 
\usepackage{amsmath}
\usepackage{pslatex}

\usepackage{graphicx}
\usepackage{wrapfig}
\usepackage{float} 

\title{The Perceptimatic English Benchmark for Speech Perception Models}

\author{{\large \bf Juliette Millet (juliette.millet@cri-paris.org)}\\
 Universit{\'{e}} de Paris, LLF, CNRS, Paris, France\\
 CoML, ENS/CNRS/EHESS/INRIA/PSL Research University, Paris, France\\ CRI, Département Frontières du Vivant et de l'Apprendre, IIFR, Universit{\'{e}} de Paris
  \AND {\large \bf Ewan Dunbar (ewan.dunbar@univ-paris-diderot.fr)} \\
 Universit{\'{e}} de Paris, LLF, CNRS, Paris, France\\
 CoML, ENS/CNRS/EHESS/INRIA/PSL Research University, Paris, France
  }

\begin{document}

\maketitle

\begin{abstract}

We present the Perceptimatic English Benchmark, an open experimental benchmark for evaluating quantitative models of speech perception in English. The benchmark consists of ABX stimuli along with the responses of 91 American English-speaking listeners. The stimuli test discrimination of a large number of English and French phonemic contrasts. They are extracted directly from corpora of read speech, making them appropriate for evaluating statistical acoustic models (such as those used in automatic speech recognition) trained on typical speech data sets. We show that phone discrimination is correlated with several types of models, and give recommendations for researchers seeking easily calculated norms of acoustic distance on experimental stimuli. We show that DeepSpeech, a standard English speech recognizer, is more specialized on English phoneme discrimination than English listeners, and is poorly correlated with their behaviour, even though it yields a low error on the decision task given to humans. 


\end{abstract}

\section{Introduction}

There is no accurate computational model of human speech perception that applies to real speech. 
Implemented speech perception models exist which take artificial phonetic or perceptual features as input and map them to recognized words  \cite{traceii,shortlistb}, use speech recognizers as a front-end to derive phonetic transcriptions \cite{spem}, or work on raw speech waveforms for extremely artificial utterances only \cite{tracei}. 
Yet, traditional automatic speech recognition systems directly analyze natural, recorded, continuous speech and decode it as a sequence of phonemes or words. 
We take the \emph{reverse engineering} approach \cite{reverseengineering} of concluding that the signal processing and machine learning tools underlying automatic speech recognition should thus provide a  starting point for a model of human speech perception.

Little is known, however, about the exact nature of the difference between the behaviour of human beings 
and that of speech processing tools developed for an applied purpose. We propose the \textbf{Perceptimatic English Benchmark} (PEB), an experimental human data set documenting a basic profile of English phone discrimination which is amenable to comparisons with a wide range of models.\footnote{All stimuli, human experimental data, analysis and processing scripts, and model results, are available at the following permanent link: [MASKED FOR REVIEW].}

We focus on a simple experiment for which typical speech recognition models could in principle give results comparable to humans, that of phone discrimination (typical speech recognition models are classifiers for sequences of phones). However, speech recognition models are trained on databases of continuous, natural speech, while typical experimental stimuli are individual phones, syllables, or words, read or synthesized in an effort to ensure that the phonetic properties being probed are audible. Such word-list type pronunciations, while clear to human listeners, are likely quite different from  the training data of standard speech recognition models. Models applied to them would be faced with the often difficult task of generalizing to a novel speech style. We first propose a more conservative test: the Perceptimatic English Benchmark is thus constructed out of snippets from a corpus of read speech---\emph{ecological} for typical models---tested as phone discrimination experiment items on  English  listeners.

To the degree made possible by the speech corpora from which the stimuli are extracted, we make the evaluation \emph{complete}, in the sense that they test discrimination of as many pairs of phones as possible, while being \emph{controlled} in several ways, notably in never comparing phones extracted from radically different phonetic contexts. To widen the benchmark, we also test French stimuli, which are unfamiliar both to human listeners and to models trained on English. Details are found in \textbf{\nameref{sec:benchmark}} below.

In this paper, we use the PEB to evaluate seven models that apply to real speech. 
We find that several models are predictive of humans. Surprisingly, a multilingual model---which is not trained to recognize English phonemes---and a short-duration acoustic event model---which is not trained to recognize phonemes at all---are far more predictive than a standard speech recognizer. We argue that the speech recognizer is too good at English phone discrimination.

\section{Perceptimatic English Benchmark}
\label{sec:benchmark}

\subsubsection{Experimental Task}
\label{sec:benchmark-abx}

We assess the perception of phones. We use an ABX discrimination task. 
Human participants hear three stimuli and are asked to identify which one of the first two stimuli (A or B) is more similar to the third (X). The experimenter always identifies a correct answer---in this case, by making A and B instances of two different phonetic categories, and X another example of one of the two. The accuracy of listeners' responses to a given \emph{triplet} (combination of specific stimuli into an A--B--X item) gives a measure of the discriminability of the categories to which A and B belong.

\subsubsection{Stimuli} \label{sec:benchmark-stimuli}

We construct triplets in which A, B, and X are each sequences of three consecutive phones extracted from running speech, where only the centre phone differs between A and B (for example, [\textipa{seIk}]--[\textipa{soUk}], [\textipa{zfA}]--[\textipa{zpA}]). We control the context in order to avoid mismatching different contextual allophones. We incorporate this context into the stimuli in order to avoid making the stimuli too short. The references, A and B, are uttered by the same speaker in order to avoid listeners' responding on the basis of speaker differences, while the probe, X, is uttered by a different speaker, to encourage listeners to focus on the linguistic content rather than acoustic detail. The delay between A and B is 500 milliseconds, and between B and X 650 milliseconds, as pilot subjects reported having difficulty recalling the reference stimuli when the delays were exactly equal.

Both English and French stimuli are extracted from the subset of the LibriVox audio book collection used as evaluation stimuli in the Zero Resource Speech Challenge (see \textbf{\nameref{sec:related-work}} below). The set of centre phones used is drawn from the phonemic transcriptions. We exclude phones (or phones in certain neutralizing contexts) which we expected might be subject to a merger, or which were sufficiently marginal that the transcriptions were unlikely to be reliable.
In total, the stimuli consist of 5202 triplets (2214 from English), making 461 distinct centre phone contrasts (212  English, 249 French), in a total of 201 distinct contexts (118  English, 83 French), with most phone comparisons appearing in three contexts each (a total of 47 English contrasts  appear in either one, two, or four contexts). The speakers used (15 English, 18 French)  have, in our assessment, pronunciations close to  standard American English/Metropolitan French. Not all phone comparisons occur, nor do all phone comparisons occur in the same contexts, or with the same set of speakers: we (native English and French listeners) selected the stimuli by hand out of the very large set of constructible triplets to maximize the phonetic similarity of the probe's centre phone to that of the correct answer, and minimize phonetic differences in the surrounding contexts. This is critical when extracting stimuli from natural speech: transcriptions are not always accurate, and a three-phone window is not sufficient to guarantee which of the many possible contextual variants each transcribed phone really corresponds to.\footnote{The full set of  English centre phones included in at least one item is $[$\textipa{\ae~A~b~d~D~eI~E~f~g~h~i I k  l  m n  N  oU p \*r  s  S  t  tS u U  v \textturnv   w  z}$]$. The full set of French phones included is [\textipa{a \~A b d e E \~E f g i j k l m n \textltailn~o \o~O \~O p K s S t u v w y z Z}]. For the full list of pairs and contexts tested, see the online repository.} Each set of three stimuli appears in four distinct items, corresponding to orders AB--A (that is, X is another instance of the three-phone sequence A), BA--B, AB--B, and BA--A.

\subsubsection{Reference Data Collection}

\label{sec:benchmark-methods}

The data set includes 91 participants reporting English as the language to which they were primarily exposed before the age of eight. They performed the task on Amazon Mechanical Turk (US participants) with the \textsc{LMEDS} software \cite{mahrt2013lmeds} and were paid for participation.\footnote{\citeNP{kleinschmidt2015robust} made a detailed comparison of data from an in-lab speech perception experiment with a Mechanical Turk replication and found a close correspondence between the results.} We asked participants to use headphones, to do the task in a quiet environment, and to check the sound volume before the experiment began. 15 additional participants were tested but did not meet the language background requirements, and 65 were rejected for failing at least three out of twelve catch trials or not finishing the task.\footnote{The catch trials consisted of additional, highly distinct three-phone  ABX stimuli, including several which required participants to distinguish \emph{cat} from \emph{dog}.}

For testing, items were counterbalanced into lists of 190 triplets per participant, such that no participant was tested twice on the same contrast, and such that the combination of speakers was not predictive of the right answer. Each stimulus was tested three times, so that most contrasts were tested at least 36 times. Participants responded as to which of the two reference stimuli the probe corresponded to on a six-point scale, ranging from  \emph{first stimuli for sure} to \emph{second stimuli for sure}, with two intermediate degrees of certainty for each reference stimulus. The benchmark includes both these responses and a binarized version, taking into account the participant's choice but not their reported certainty. Here we report only analysis of the binarized responses to avoid questions about how to model participants' use of the scale (preliminary analyses on the scaled responses indicate that the results are qualitatively the same).

\section{Generating Model Predictions}

\label{sec:abx-prediction}

For each experimental stimulus, we suppose that we can apply a model to the audio file and extract that model's representation of the stimulus (see \label{sec:experiments} below for examples). To predict human responses, we compute distances $d(\mbox{Target},\mbox{X})$, between the probe and the correct matching stimulus, and $d(\mbox{Other},\mbox{X})$, between the probe and the other reference stimulus, to generate a degree of correct discriminability $\delta = d(\mbox{Other},\mbox{X}) - d(\mbox{Target},\mbox{X})$. If $\delta>0$, then the model treats the probe as being more similar to the correct than the incorrect answer. Our goal is to assess whether humans' perceived similarity matches the model's distances. Humans' responses are stochastic, and need not use a  threshold at the point of maximal perceived similarity. This leads us to use  a binomial generalized linear model with an intercept parameter.

This is not the only possible linking hypothesis, but it is broadly applicable, and allows for a distance function to be selected that is appropriate to the type of representation being tested. All the models we consider in this paper yield representations of variable length (they output vector sequences---one vector per time frame---and the stimuli are not all of the same duration). Thus, we use distance functions based on dynamic time warping. Dynamic time warping takes two  sequences $C$ and $D$ as input, as well as a function $\gamma$ for comparing pairs of sequence elements. It aligns  $C$ and $D$ by matching the elements of one to the other so as to minimize the sum of $\gamma(c, d)$ for all matched elements $(c,d)$. Each element of $C$ must to be matched with at least one element of $D$, and alignments must respect temporal order. Here we calculate distances between stimuli $C = {c_1, c_2,...c_p}$ and $D = {d_1, d_2, ... d_q}$ as:
\begin{align}
    d(C,D) = \frac{\sum_{c_i, d_j \text{ are matched }} \gamma(c_i, d_j)}{max(p, q)}
\end{align}
For the models analyzed here, we take $\gamma$ to be either a symmetrised Kullblack--Leibler divergence\footnote{We replace zero elements with a very small constant to avoid division by zero.} (for models that output probability vectors), or a cosine distance. Where $\mathbf{x}$ and $\mathbf{y}$ are $N$-dimensional vectors, they are defined as:
\begin{align}
    \gamma_{KL}(\mathbf{x},\mathbf{y}) = \frac{1}{2}\left[\sum_{i=1}^{N} x_i \log(\frac{x_i}{y_i}) + \sum_{i=1}^{N} y_i \log(\frac{y_i}{x_i})\right]
    \label{eq:kl}
\end{align}
\begin{align}
    \gamma_{cos}(\mathbf{x},\mathbf{y}) = \frac{1}{\pi}\arccos\left(\frac{\sum_{i=1}^{N} x_iy_i}{ \sqrt{\sum_{i=1}^{N} x_i^2} \sqrt{\sum_{i=1}^{N} y_i^2}}\right)
    \label{eq:cosine}
\end{align}

\section{Experiments}

\label{sec:experiments}

We apply the methods described in the previous section. Unless stated, we take $\gamma$ to be the cosine distance \eqref{eq:cosine}.


\subsubsection{Dirichlet Process Gaussian Mixture Model}

We evaluate a Dirichlet process Gaussian mixture model (DPGMM) as proposed by \citeA{chen2015parallel}. Given a training set of speech recordings in a language, the model performs non-parametric Bayesian clustering on the entire database, treated as an unordered collection of instantaneous acoustic feature vectors (see \textbf{\nameref{sec:mfcc}} below). It models short-duration acoustic events. A fitted model consists of an optimal set of Gaussian distributions---typically several hundred. The model thus preserves fine-grained temporal and acoustic detail, while still modelling a specific language. It does not use phoneme labels.  
Passing over a new sample at a fixed analysis rate (in our case, analyzing 25 milliseconds of signal every 10 milliseconds), each instant of signal is mapped to a vector of posterior probabilities over the Gaussians in the model. We take $\gamma$ to be the symmetrized KL divergence \eqref{eq:kl}. We apply the English model described in \citeA{millet2019comparing}, trained on 34 hours 
of English speech taken from the LibriVox dataset (no overlap with the stimuli or speakers in Perceptimatic). 

\subsubsection{Bottleneck features}

We evaluate three models proposed in \citeA{bottleneck}. These \emph{bottleneck} models are trained to label speech with \emph{phone states}. Phone states are temporal analysis units used by certain speech recognizers: each phone of the language is modelled as having (in the typical three-state model) a beginning, middle, and end state, each with different acoustic properties. The bottleneck models are trained on speech data labeled annotated with an attribution to phone states.
They are neural networks trained to predict the phone state associated with a given instant of speech, on the basis of its acoustic features, accompanied by 310 ms of surrounding context. This model is thus  optimized to predict a slightly more temporally fine-grained version of standard phoneme labels.
``Bottleneck'' 
refers to a hidden layer that has significantly lower dimension than the 
other layers. The features we use are the contents of this layer, for each instant of signal.  
We evaluate \emph{English monophone,} \emph{English triphone,} and \emph{multilingual} models.\footnote{Referred to by \citeA{bottleneck} as \emph{FisherMono,} \emph{FisherTri,} and \emph{BabelMulti}.} The English monophone model is optimized to predict states for English phonemes. The English triphone model is optimized to predict states for contextual allophones. The multilingual model is trained on data from seventeen phonetically diverse languages (not including English), optimized to label phoneme states in any of these languages (if the same sound belongs to different inventories, it is treated as distinct, for a total of 1032 possible phonemes).  

\subsubsection{DeepSpeech}

DeepSpeech \cite{hannun2014deep} is 
a neural 
automatic speech recognition model used in the Mozilla speech tools.
The model uses bi-directional recurrent units, which integrate information both forwards and backwards in time, to predict text transcriptions (sequences of letters, not phones) from speech. We can examine the state of any of its several internal layers corresponding to any instant of signal. After scoring each layer on its performance on the (artificial) phone discrimination evaluation described in \citeA{dunbar2017zero}, we found that layer five was optimal. We thus analyze the outputs from that layer. The model has a training objective related to that of the English bottleneck models (predicting text), but the recurrent units allow it to model long distance temporal dependencies. 
We use Mozilla DeepSpeech 0.4.1\footnote{\url{https://github.com/mozilla/DeepSpeech/releases/tag/v0.4.1}}, which is trained on  the Fisher \cite{fisher} and Switchboard \cite{godfrey1992switchboard} telephone corpora and  the LibriSpeech audio book corpus \cite{panayotov2015librispeech}. The model achieves an 8.26\% word error rate on the LibriSpeech clean test evaluation.


\subsubsection{Articulatory Reconstruction}

To explore whether similarities at the level of articulation are more predictive of  humans' behaviour, we evaluate a neural \emph{articulatory reconstruction} model \cite{parrot2019independent}, trained to reconstruct continuous electromagnetic articulography (EMA) coil position trajectories from speech recordings (tongue body, tongue tip, tongue dorsum, upper lip, lower lip, lower incisor). The model is trained on the EMA-IEEE corpus \cite{Haskins}, approximately six hours of read English speech, paired with EMA recordings, from eight speakers.

\subsubsection{Mel Filterbank Cepstral Coefficients} \label{sec:mfcc}

We use Kaldi \cite{povey2011kaldi} to extract 
13 Mel filterbank cepstral coefficients (MFCC): one vector every 10 milliseconds, each analyzing 25 milliseconds of signal. These  audio representations, used standardly as input to speech recognition, are the result of  a low-resolution spectral analysis and  a discrete cosine transformation.  We add the first and second derivatives, for a total of 39 dimensions, and apply 
mean-variance normalization over a moving three-second window. This approach, like the multilingual bottleneck features, does not specifically model English; unlike that model, this is a fixed transformation, not tuned to any language, or indeed to speech at all.

\section{Results}

\label{sec:results}

\subsubsection{Performance on the Experimental Task}

We compute the mean accuracies\footnote{Scoring accuracy first by stimulus, then averaging by contrast, then overall.} for each of the models, scoring stimuli as correct where $\delta >0$.
The results in Table \ref{tab:results} indicate that the models' performance is generally better than the human listeners in the PEB. This implies that, to the extent that any of these models accurately captures listeners' perceived discriminability, listeners' behaviour on the task, unsurprisingly, cannot correspond to a hard decision at the optimal decision threshhold. The results also indicate, as expected, a small native language effect---a decrease in listeners' discrimination accuracy for the non-English stimuli. Such an effect is also captured by all the models trained on English. We observe that some models show native language effects numerically much larger than human listeners, a point we return to below.


\begin{table}[]
\begin{center}

\begin{tabular}{l|p{16pt}|p{16pt}p{9pt}p{17pt}p{16pt}p{12pt}p{8pt}p{18pt}}
               & PEB & GMM & DS   & BEnM & BEnT & BMu  & Art  & MFCC \\ \hline
En & 79.5      & 88.3  & 89.5 & 91.2  & 90.3  & 88.9 & 77.3 & 78.6 \\
Fr  & 76.7      & 82.0  & 80.2 & 87.6  & 88.8  & 88.5 & 70.1 & 78.3
\end{tabular}

\end{center}
\caption{Percent accuracies for humans (PEB) and models (the bigger the better). GMM is for DPGMM, DS for DeepSpeech. BEnM, BEnT and BMu are (in order) for monophone English, triphone English  and multilingual bottleneck models. Art is for articulatory reconstruction.}
\label{tab:results}
\end{table}


\subsubsection{Prediction}

In order to see which model best predicts the human results,\footnote{Here we report results on both the English (native) and French (non-native) stimuli. In the interest of analyzing stimuli that are maximally ecological for the models tested, we also analyzed the results of the native-language perception task only. The results are qualitatively identical, so we omit them in the interest of space. The table is available in the online repository.} we fit probit regression models with a coefficient for the $\delta$ discriminability score corresponding to the given model. The dependent variable is whether the trial response was correct (1: accurate,  0: inaccurate). 
To correct for effects that are not of interest, 
the models each also include a coefficient for whether the correct answer was A or B, a coefficient for the position of the trial in the experimental list, and a coefficient for participant. 

We use differences in log likelihood for model comparison, obtaining confidence intervals by repeatedly drawing balanced subsamples ($N=43358$):
for each stimulus, we draw three observations  without replacement. 
The results, in Table \ref{tab:conf_inter}, show that the most predictive approaches are short-term acoustic event modelling (DPGMM) and bottleneck phone state predictors, with the English monophone (phoneme) predictor model showing non-significantly poorer performance than the allophonic and multilingual ones. 

\begin{table}[]
\begin{center}

\begin{tabular}{l|p{12pt}p{20pt}p{24pt}p{20pt}p{12pt}p{12pt}}
      & BMu       & BEngT & BEngM & MFCC         & DS           & Art          \\ \hline
GMM & 3         & 9     & 28    & \textbf{204} & \textbf{249} & \textbf{257} \\
BMu   &           & 6     & 24    & \textbf{202} & \textbf{246} & \textbf{254} \\
BEngT & \textbf{} &       & 19    & \textbf{196} & \textbf{241} & \textbf{248} \\
BEngM &           &       &       & \textbf{177} & \textbf{222} & \textbf{229} \\
MFCC  &           &       &       &              & \textbf{45}  & \textbf{52}  \\
DS    &           &       &       &              &              & 8           
\end{tabular}

\end{center}
\caption{Mean of resampled differences in log likelihood between models. Models are ordered by column and be row in descending order of their performance, with better models on the left/top. Positive numbers indicate that the model indicated in the given row is better than the model indicated in the column. Bolded results have 95\% confidence intervals that exclude zero. GMM is for DPGMM, DS for DeepSpeech. BEnM, BEnT and BMu are (in order) for monophone English, triphone English  and multilingual bottleneck models. Art is for articulatory reconstruction.
}
\label{tab:conf_inter}
\end{table}

\section{Discussion}

\label{sec:discussion}

The results of our English phone discrimination benchmark are best predicted by the DPGMM's short-duration acoustic event modelling and the three bottleneck phone state classification models, consistent with \citeA{millet2019comparing} and \citeA{nikamemoire}.  These do substantially better than  generic audio features.
Two of the bottleneck models are trained to predict English phoneme/allophone labels, but the multilingual model is not, which makes its performance all the more surprising. Neither is the DPGMM model, which, although trained on English, models 25 millisecond acoustic events into combinations of hundreds of detailed acoustic categories, and is thus much more fine-grained than typical phonetic annotation.

The articulatory reconstruction model is not very predictive of human behaviour. The likely reasons are  simple. First, predicting articulatory parameters for novel speakers is difficult, and this model is far from  state-of-the-art performance. Second, the model does not predict a complete set of articulators. It is thus unsurprising that, when scored on the experimental task, this model is worse than even the acoustic features.

The continuous speech recognizer (DeepSpeech) is also bad at predicting human behaviour, but, unlike the articulatory reconstruction, performs well on the experimental task. This model is different from the English bottleneck models in two ways. First, it is in principle capable of taking into account longer temporal dependencies than the finite 310 ms window used by the bottleneck model.  Second, it is optimized not to predict phones, but orthographic (letter) transcriptions. These are quite similar, but English orthography is still not completely transparent, which might help explain the model's behaviour: distinct sequences of phones correspond to distinct sequences of letters (thus, allow for a high score on the experimental task), but the representation's distances may capture similarities and differences exclusively found in spelling. We also note, however, that the model shows the largest discrepancy between the English  and  French stimuli (larger than the English listeners'). It  is not immediately obvious how this could be attributed to predicting English letters versus phones.

\begin{figure}
\begin{center}
\includegraphics[width=0.5\textwidth]{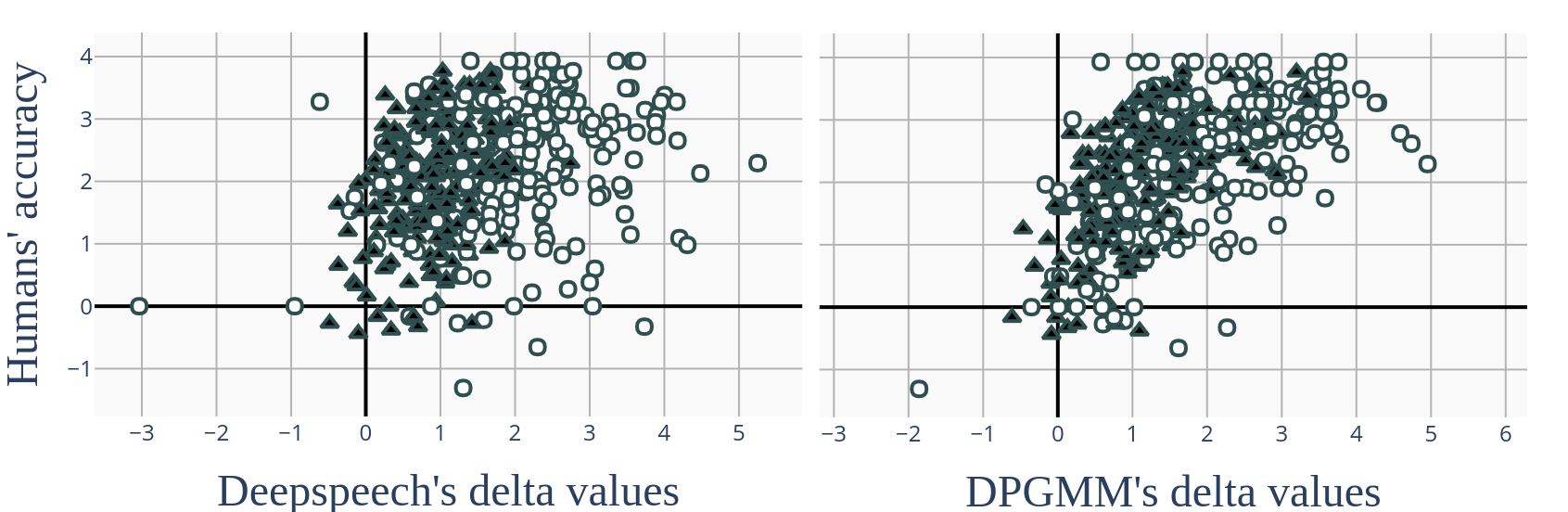}

\caption{Normalized human accuracy versus normalized $\delta$ for (left) the DeepSpeech speech recognizer and (right) the DPGMM short-duration acoustic model. White circles are English stimuli and black triangles are French stimuli. 
}
\label{fig:compa}
\end{center}

\end{figure}

This difference is clear from Figure \ref{fig:compa} (left), which plots DeepSpeech's $\delta$ discriminability scores against listeners' averaged accuracy for each contrast, colour-coded for whether the items are English or  French. We observe a clear separation in the distributions of the DeepSpeech's discriminability of English contrasts (concentrated on the right-hand part of the graph, where the model is better) versus French contrasts. This separation is not visually salient in humans, nor in the the DPGMM model (right). 
This model seems to be over-trained on the task of discriminating English phones.  

Finally, we consider the nature of the benchmark itself. While speaker variability was introduced in order to prevent listeners from attending to acoustic details, the delay between stimuli is still relatively short, meaning that listeners need not rely heavily on memory, and will thus still have reasonable access to detail. The stimuli are also short, and often do not correspond to full syllables, so that listeners may not treat them as fully speech-like. The fact that the A and B stimuli are  from the same speaker may also attune speakes to small differences between those two stimuli, potentially thus attuning them to low-level differences overall. If listeners focus on detail, then the fact that the drop in human performance on non-native stimuli is small is unsurprising. The fact that the multilingual and DPGMM models are good at predicting the behaviour of English-speaking listeners may prove to be a consequence of this particular mode of listening. Other benchmark tasks are needed to obtain a fuller picture.

Stimuli extracted from running speech may be more ecological for evaluating typical speech recognition models, but they are difficult to interpret. While context was, in principle, held constant across each stimulus triplet, in reality, it is very difficult to get phonetically well-matched contexts in natural speech. Although the stimuli were selected by hand to minimize the differences due to surrounding context, they are not perfectly controlled, which means that the target (centre) phone is not the only thing driving human listeners' decisions. Among the more difficult English contrasts for listeners here are English [f]--[v], which should not be particularly difficult, and French [f]--[y], which should be trivially easy. Items like these evidently do not highlight the desired contrast---and the fact that the locus of contrast was not always apparent might also have led listeners to attend to acoustic detail.

\section{Related work}

\label{sec:related-work}

Our data set 
is 
in the spirit of other cognitive benchmarks for artificial intelligence (syntax: \citeNP{blimp}; intuitive physics: \citeNP{intphys}; question answering: \citeNP{natquest}).
In speech perception, the idea of matching human behaviour is not new  \cite{kleinschmidt2015robust,feldman2007rational,schatz2019early,schatz2017asr,thomashmm}, and is an echo of the literature on modelling phonetic learning, most notably \citeA{guenthergjaja}, who qualitatively compared their modelled distances to similarities reported in the literature for human listeners.
To our knowledge, the only previous work providing stimuli, human responses, and  recommendations for generating predictions at the individual stimulus level with a wide range of  models is \citeA{millet2019comparing}. Those stimuli only tested a narrow range of cross-linguistic phone contrasts, however, and were non-words read in a word-list style, rather than extracts of natural, running speech. 

The PEB stimuli are drawn from the evaluation for the Zero Resource Speech Challenge 2017 \cite{dunbar2017zero}, widely used in evaluating  unsupervised speech models. The PEB complements  this existing  measure (not scored against humans), and can be applied to any model tested on it.

\section{Summary of Contributions}

\label{sec:conclusions}

We have presented the \textbf{Perceptimatic English Benchmark}, an open English-language benchmark for computational models of human speech perception made up of stimuli that are ecological for typical speech models. It is the only  open data set we know of that systematically probes a wide range of phone contrasts and is easy to compare with computational models. We have shown, for the first time, that a standard speech recognizer is not predictive of human phone classification behaviour, while models not optimized to recognize English phonemes are  (a quasi-universal phone classifier and a model of short-duration acoustic events). The multilingual model is easy to use off-the-shelf,\footnote{\url{https://coml.lscp.ens.fr/docs/shennong/}.} and we recommend it to researchers needing an estimate of perceptual distance.

\bibliographystyle{apacite}

\setlength{\bibleftmargin}{.125in}
\setlength{\bibindent}{-\bibleftmargin}

\bibliography{main}

\begin{thebibliography}{}

\bibitem [\protect \citeauthoryear {%
Chen%
, Leung%
, Xie%
, Ma%
\BCBL {}\ \BBA {} Li%
}{%
Chen%
\ \protect \BOthers {.}}{%
{\protect \APACyear {2015}}%
}]{%
chen2015parallel}
\APACinsertmetastar {%
chen2015parallel}%
\begin{APACrefauthors}%
Chen, H.%
, Leung, C\BHBI C.%
, Xie, L.%
, Ma, B.%
\BCBL {}\ \BBA {} Li, H.%
\end{APACrefauthors}%
\unskip\
\newblock
\APACrefYearMonthDay{2015}{}{}.
\newblock
{\BBOQ}\APACrefatitle {{Parallel inference of Dirichlet process Gaussian
  mixture models for unsupervised acoustic modeling: A feasibility study}}
  {{Parallel inference of Dirichlet process Gaussian mixture models for
  unsupervised acoustic modeling: A feasibility study}}.{\BBCQ}
\newblock
\BIn{} \APACrefbtitle {{INTERSPEECH-16}.} {{INTERSPEECH-16}.}
\PrintBackRefs{\CurrentBib}

\bibitem [\protect \citeauthoryear {%
Cieri%
, Miller%
\BCBL {}\ \BBA {} Walker%
}{%
Cieri%
\ \protect \BOthers {.}}{%
{\protect \APACyear {2004}}%
}]{%
fisher}
\APACinsertmetastar {%
fisher}%
\begin{APACrefauthors}%
Cieri, C.%
, Miller, D.%
\BCBL {}\ \BBA {} Walker, K.%
\end{APACrefauthors}%
\unskip\
\newblock
\APACrefYearMonthDay{2004}{}{}.
\newblock
{\BBOQ}\APACrefatitle {The Fisher Corpus: a Resource for the Next Generations
  of Speech-to-Text} {The fisher corpus: a resource for the next generations of
  speech-to-text}.{\BBCQ}
\newblock
\BIn{} \APACrefbtitle {LREC.} {Lrec.}
\PrintBackRefs{\CurrentBib}

\bibitem [\protect \citeauthoryear {%
Dunbar%
\ \protect \BOthers {.}}{%
Dunbar%
\ \protect \BOthers {.}}{%
{\protect \APACyear {2017}}%
}]{%
dunbar2017zero}
\APACinsertmetastar {%
dunbar2017zero}%
\begin{APACrefauthors}%
Dunbar, E.%
, Cao, X\BPBI N.%
, Benjumea, J.%
, Karadayi, J.%
, Bernard, M.%
, Besacier, L.%
\BDBL {}Dupoux, E.%
\end{APACrefauthors}%
\unskip\
\newblock
\APACrefYearMonthDay{2017}{}{}.
\newblock
{\BBOQ}\APACrefatitle {The {Z}ero {R}esource {S}peech {C}hallenge 2017} {The
  {Z}ero {R}esource {S}peech {C}hallenge 2017}.{\BBCQ}
\newblock
\BIn{} \APACrefbtitle {{2017 {IEEE} Workshop on Automatic Speech Recognition
  and Understanding (ASRU)}} {{2017 {IEEE} Workshop on Automatic Speech
  Recognition and Understanding (ASRU)}}\ (\BPGS\ 323--330).
\PrintBackRefs{\CurrentBib}

\bibitem [\protect \citeauthoryear {%
Dupoux%
}{%
Dupoux%
}{%
{\protect \APACyear {2018}}%
}]{%
reverseengineering}
\APACinsertmetastar {%
reverseengineering}%
\begin{APACrefauthors}%
Dupoux, E.%
\end{APACrefauthors}%
\unskip\
\newblock
\APACrefYearMonthDay{2018}{}{}.
\newblock
{\BBOQ}\APACrefatitle {Cognitive science in the era of artificial intelligence:
  A roadmap for reverse-engineering the infant language-learner} {Cognitive
  science in the era of artificial intelligence: A roadmap for
  reverse-engineering the infant language-learner}.{\BBCQ}
\newblock
\APACjournalVolNumPages{Cognition}{173}{}{43--59}.
\PrintBackRefs{\CurrentBib}

\bibitem [\protect \citeauthoryear {%
Elman%
\ \BBA {} McClelland%
}{%
Elman%
\ \BBA {} McClelland%
}{%
{\protect \APACyear {2015}}%
}]{%
tracei}
\APACinsertmetastar {%
tracei}%
\begin{APACrefauthors}%
Elman, J.%
\BCBT {}\ \BBA {} McClelland, J.%
\end{APACrefauthors}%
\unskip\
\newblock
\APACrefYearMonthDay{2015}{}{}.
\newblock
{\BBOQ}\APACrefatitle {Exploiting the lawful variability in the speech wave}
  {Exploiting the lawful variability in the speech wave}.{\BBCQ}
\newblock
\BIn{} J.~Perkell\ \BBA {} D.~Klatt\ (\BEDS), (\BVOL~335, \BPGS\ 71--90).
\newblock
\APACaddressPublisher{Hillsdale, NJ}{Erlbaum}.
\PrintBackRefs{\CurrentBib}

\bibitem [\protect \citeauthoryear {%
Feldman%
\ \BBA {} Griffiths%
}{%
Feldman%
\ \BBA {} Griffiths%
}{%
{\protect \APACyear {2007}}%
}]{%
feldman2007rational}
\APACinsertmetastar {%
feldman2007rational}%
\begin{APACrefauthors}%
Feldman, N\BPBI H.%
\BCBT {}\ \BBA {} Griffiths, T\BPBI L.%
\end{APACrefauthors}%
\unskip\
\newblock
\APACrefYearMonthDay{2007}{}{}.
\newblock
{\BBOQ}\APACrefatitle {A rational account of the perceptual magnet effect} {A
  rational account of the perceptual magnet effect}.{\BBCQ}
\newblock
\BIn{} \APACrefbtitle {Proceedings of the Annual Meeting of the Cognitive
  Science Society} {Proceedings of the annual meeting of the cognitive science
  society}\ (\BVOL~29).
\PrintBackRefs{\CurrentBib}

\bibitem [\protect \citeauthoryear {%
Fer%
\ \protect \BOthers {.}}{%
Fer%
\ \protect \BOthers {.}}{%
{\protect \APACyear {2017}}%
}]{%
bottleneck}
\APACinsertmetastar {%
bottleneck}%
\begin{APACrefauthors}%
Fer, R.%
, Matejka, P.%
, Grezl, F.%
, Plchot, O.%
, Vesely, K.%
\BCBL {}\ \BBA {} Cernocky, J\BPBI H.%
\end{APACrefauthors}%
\unskip\
\newblock
\APACrefYearMonthDay{2017}{}{}.
\newblock
{\BBOQ}\APACrefatitle {Multilingually trained bottleneck features in spoken
  language recognition} {Multilingually trained bottleneck features in spoken
  language recognition}.{\BBCQ}
\newblock
\APACjournalVolNumPages{Computer Speech and Language}{46}{Supplement C}{252 -
  267}.
\PrintBackRefs{\CurrentBib}

\bibitem [\protect \citeauthoryear {%
Godfrey%
, Holliman%
\BCBL {}\ \BBA {} McDaniel%
}{%
Godfrey%
\ \protect \BOthers {.}}{%
{\protect \APACyear {1992}}%
}]{%
godfrey1992switchboard}
\APACinsertmetastar {%
godfrey1992switchboard}%
\begin{APACrefauthors}%
Godfrey, J\BPBI J.%
, Holliman, E\BPBI C.%
\BCBL {}\ \BBA {} McDaniel, J.%
\end{APACrefauthors}%
\unskip\
\newblock
\APACrefYearMonthDay{1992}{}{}.
\newblock
{\BBOQ}\APACrefatitle {SWITCHBOARD: Telephone speech corpus for research and
  development} {Switchboard: Telephone speech corpus for research and
  development}.{\BBCQ}
\newblock
\BIn{} \APACrefbtitle {[Proceedings] ICASSP-92: 1992 IEEE International
  Conference on Acoustics, Speech, and Signal Processing} {[proceedings]
  icassp-92: 1992 ieee international conference on acoustics, speech, and
  signal processing}\ (\BVOL~1, \BPGS\ 517--520).
\PrintBackRefs{\CurrentBib}

\bibitem [\protect \citeauthoryear {%
Guenther%
\ \BBA {} Gjaja%
}{%
Guenther%
\ \BBA {} Gjaja%
}{%
{\protect \APACyear {1996}}%
}]{%
guenthergjaja}
\APACinsertmetastar {%
guenthergjaja}%
\begin{APACrefauthors}%
Guenther, F.%
\BCBT {}\ \BBA {} Gjaja, M.%
\end{APACrefauthors}%
\unskip\
\newblock
\APACrefYearMonthDay{1996}{}{}.
\newblock
{\BBOQ}\APACrefatitle {The perceptual magnet effect as an emergent property of
  neural map formation} {The perceptual magnet effect as an emergent property
  of neural map formation}.{\BBCQ}
\newblock
\APACjournalVolNumPages{The Journal of the Acoustical Society of
  America}{100}{2}{1111--1121}.
\PrintBackRefs{\CurrentBib}

\bibitem [\protect \citeauthoryear {%
Hannun%
\ \protect \BOthers {.}}{%
Hannun%
\ \protect \BOthers {.}}{%
{\protect \APACyear {2014}}%
}]{%
hannun2014deep}
\APACinsertmetastar {%
hannun2014deep}%
\begin{APACrefauthors}%
Hannun, A.%
, Case, C.%
, Casper, J.%
, Catanzaro, B.%
, Diamos, G.%
, Elsen, E.%
\BDBL {}others%
\end{APACrefauthors}%
\unskip\
\newblock
\APACrefYearMonthDay{2014}{}{}.
\newblock
{\BBOQ}\APACrefatitle {Deep speech: Scaling up end-to-end speech recognition}
  {Deep speech: Scaling up end-to-end speech recognition}.{\BBCQ}
\newblock
\APACjournalVolNumPages{arXiv preprint arXiv:1412.5567}{}{}{}.
\PrintBackRefs{\CurrentBib}

\bibitem [\protect \citeauthoryear {%
Jurov%
}{%
Jurov%
}{%
{\protect \APACyear {2019}}%
}]{%
nikamemoire}
\APACinsertmetastar {%
nikamemoire}%
\begin{APACrefauthors}%
Jurov, N.%
\end{APACrefauthors}%
\unskip\
\newblock
\APACrefYear{2019}.
\newblock
\APACrefbtitle {{Phonetics or Phonology? Modelling Non-Native Perception}}
  {{Phonetics or Phonology? Modelling Non-Native Perception}}.
\newblock
\BUMTh, Université Paris Diderot, Paris, France.
\PrintBackRefs{\CurrentBib}

\bibitem [\protect \citeauthoryear {%
Kleinschmidt%
\ \BBA {} Jaeger%
}{%
Kleinschmidt%
\ \BBA {} Jaeger%
}{%
{\protect \APACyear {2015}}%
}]{%
kleinschmidt2015robust}
\APACinsertmetastar {%
kleinschmidt2015robust}%
\begin{APACrefauthors}%
Kleinschmidt, D.%
\BCBT {}\ \BBA {} Jaeger, T\BPBI F.%
\end{APACrefauthors}%
\unskip\
\newblock
\APACrefYearMonthDay{2015}{}{}.
\newblock
{\BBOQ}\APACrefatitle {{Robust speech perception: Recognize the familiar,
  generalize to the similar, and adapt to the novel}} {{Robust speech
  perception: Recognize the familiar, generalize to the similar, and adapt to
  the novel}}.{\BBCQ}
\newblock
\APACjournalVolNumPages{Psychological Review}{122}{2}{148--203}.
\PrintBackRefs{\CurrentBib}

\bibitem [\protect \citeauthoryear {%
Kwiatkowski%
\ \protect \BOthers {.}}{%
Kwiatkowski%
\ \protect \BOthers {.}}{%
{\protect \APACyear {2019}}%
}]{%
natquest}
\APACinsertmetastar {%
natquest}%
\begin{APACrefauthors}%
Kwiatkowski, T.%
, Palomaki, J.%
, Redfield, O.%
, Collins, M.%
, Parikh, A.%
, Alberti, C.%
\BDBL {}others%
\end{APACrefauthors}%
\unskip\
\newblock
\APACrefYearMonthDay{2019}{}{}.
\newblock
{\BBOQ}\APACrefatitle {Natural questions: a benchmark for question answering
  research} {Natural questions: a benchmark for question answering
  research}.{\BBCQ}
\newblock
\APACjournalVolNumPages{Transactions of the Association for Computational
  Linguistics}{7}{}{453--466}.
\PrintBackRefs{\CurrentBib}

\bibitem [\protect \citeauthoryear {%
Mahrt%
}{%
Mahrt%
}{%
{\protect \APACyear {2016}}%
}]{%
mahrt2013lmeds}
\APACinsertmetastar {%
mahrt2013lmeds}%
\begin{APACrefauthors}%
Mahrt, T.%
\end{APACrefauthors}%
\unskip\
\newblock
\APACrefYearMonthDay{2016}{}{}.
\newblock
\APACrefbtitle {{LMEDS: Language markup and experimental design software}.}
  {{LMEDS: Language markup and experimental design software}.}
\PrintBackRefs{\CurrentBib}

\bibitem [\protect \citeauthoryear {%
McClelland%
\ \BBA {} Elman%
}{%
McClelland%
\ \BBA {} Elman%
}{%
{\protect \APACyear {1986}}%
}]{%
traceii}
\APACinsertmetastar {%
traceii}%
\begin{APACrefauthors}%
McClelland, J.%
\BCBT {}\ \BBA {} Elman, J.%
\end{APACrefauthors}%
\unskip\
\newblock
\APACrefYearMonthDay{1986}{}{}.
\newblock
{\BBOQ}\APACrefatitle {{Interactive processes in speech perception: The TRACE
  model}} {{Interactive processes in speech perception: The TRACE
  model}}.{\BBCQ}
\newblock
\APACjournalVolNumPages{Cognitive Psychology}{18}{}{1--86}.
\PrintBackRefs{\CurrentBib}

\bibitem [\protect \citeauthoryear {%
Millet%
, Jurov%
\BCBL {}\ \BBA {} Dunbar%
}{%
Millet%
\ \protect \BOthers {.}}{%
{\protect \APACyear {2019}}%
}]{%
millet2019comparing}
\APACinsertmetastar {%
millet2019comparing}%
\begin{APACrefauthors}%
Millet, J.%
, Jurov, N.%
\BCBL {}\ \BBA {} Dunbar, E.%
\end{APACrefauthors}%
\unskip\
\newblock
\APACrefYearMonthDay{2019}{}{}.
\newblock
{\BBOQ}\APACrefatitle {Comparing unsupervised speech learning directly to human
  performance in speech perception} {Comparing unsupervised speech learning
  directly to human performance in speech perception}.{\BBCQ}.
\PrintBackRefs{\CurrentBib}

\bibitem [\protect \citeauthoryear {%
Norris%
\ \BBA {} McQueen%
}{%
Norris%
\ \BBA {} McQueen%
}{%
{\protect \APACyear {2008}}%
}]{%
shortlistb}
\APACinsertmetastar {%
shortlistb}%
\begin{APACrefauthors}%
Norris, D.%
\BCBT {}\ \BBA {} McQueen, J.%
\end{APACrefauthors}%
\unskip\
\newblock
\APACrefYearMonthDay{2008}{}{}.
\newblock
{\BBOQ}\APACrefatitle {{Shortlist B: a Bayesian model of continuous speech
  recognition}} {{Shortlist B: a Bayesian model of continuous speech
  recognition}}.{\BBCQ}
\newblock
\APACjournalVolNumPages{Psychological Review}{115}{2}{357--395}.
\PrintBackRefs{\CurrentBib}

\bibitem [\protect \citeauthoryear {%
Panayotov%
, Chen%
, Povey%
\BCBL {}\ \BBA {} Khudanpur%
}{%
Panayotov%
\ \protect \BOthers {.}}{%
{\protect \APACyear {2015}}%
}]{%
panayotov2015librispeech}
\APACinsertmetastar {%
panayotov2015librispeech}%
\begin{APACrefauthors}%
Panayotov, V.%
, Chen, G.%
, Povey, D.%
\BCBL {}\ \BBA {} Khudanpur, S.%
\end{APACrefauthors}%
\unskip\
\newblock
\APACrefYearMonthDay{2015}{}{}.
\newblock
{\BBOQ}\APACrefatitle {Librispeech: an asr corpus based on public domain audio
  books} {Librispeech: an asr corpus based on public domain audio
  books}.{\BBCQ}
\newblock
\BIn{} \APACrefbtitle {2015 IEEE International Conference on Acoustics, Speech
  and Signal Processing (ICASSP)} {2015 ieee international conference on
  acoustics, speech and signal processing (icassp)}\ (\BPGS\ 5206--5210).
\PrintBackRefs{\CurrentBib}

\bibitem [\protect \citeauthoryear {%
Parrot%
, Millet%
\BCBL {}\ \BBA {} Dunbar%
}{%
Parrot%
\ \protect \BOthers {.}}{%
{\protect \APACyear {2019}}%
}]{%
parrot2019independent}
\APACinsertmetastar {%
parrot2019independent}%
\begin{APACrefauthors}%
Parrot, M.%
, Millet, J.%
\BCBL {}\ \BBA {} Dunbar, E.%
\end{APACrefauthors}%
\unskip\
\newblock
\APACrefYearMonthDay{2019}{}{}.
\newblock
{\BBOQ}\APACrefatitle {Independent and automatic evaluation of
  acoustic-to-articulatory inversion models} {Independent and automatic
  evaluation of acoustic-to-articulatory inversion models}.{\BBCQ}
\newblock
\APACjournalVolNumPages{arXiv}{}{}{arXiv--1911}.
\PrintBackRefs{\CurrentBib}

\bibitem [\protect \citeauthoryear {%
Povey%
\ \protect \BOthers {.}}{%
Povey%
\ \protect \BOthers {.}}{%
{\protect \APACyear {2011}}%
}]{%
povey2011kaldi}
\APACinsertmetastar {%
povey2011kaldi}%
\begin{APACrefauthors}%
Povey, D.%
, Ghoshal, A.%
, Boulianne, G.%
, Burget, L.%
, Glembek, O.%
, Goel, N.%
\BDBL {}others%
\end{APACrefauthors}%
\unskip\
\newblock
\APACrefYearMonthDay{2011}{}{}.
\newblock
{\BBOQ}\APACrefatitle {The {Kaldi} speech recognition toolkit} {The {Kaldi}
  speech recognition toolkit}.{\BBCQ}
\newblock
\BIn{} \APACrefbtitle {{IEEE 2011 Workshop on Automatic Speech Recognition and
  Understanding (ASRU)}.} {{IEEE 2011 Workshop on Automatic Speech Recognition
  and Understanding (ASRU)}.}
\PrintBackRefs{\CurrentBib}

\bibitem [\protect \citeauthoryear {%
Riochet%
\ \protect \BOthers {.}}{%
Riochet%
\ \protect \BOthers {.}}{%
{\protect \APACyear {2018}}%
}]{%
intphys}
\APACinsertmetastar {%
intphys}%
\begin{APACrefauthors}%
Riochet, R.%
, Castro, M\BPBI Y.%
, Bernard, M.%
, Lerer, A.%
, Fergus, R.%
, Izard, V.%
\BCBL {}\ \BBA {} Dupoux, E.%
\end{APACrefauthors}%
\unskip\
\newblock
\APACrefYearMonthDay{2018}{}{}.
\newblock
{\BBOQ}\APACrefatitle {Intphys: A framework and benchmark for visual intuitive
  physics reasoning} {Intphys: A framework and benchmark for visual intuitive
  physics reasoning}.{\BBCQ}
\newblock
\APACjournalVolNumPages{arXiv preprint arXiv:1803.07616}{}{}{}.
\PrintBackRefs{\CurrentBib}

\bibitem [\protect \citeauthoryear {%
Scharenborg%
, Norris%
, ten Bosch%
\BCBL {}\ \BBA {} McQueen%
}{%
Scharenborg%
\ \protect \BOthers {.}}{%
{\protect \APACyear {2005}}%
}]{%
spem}
\APACinsertmetastar {%
spem}%
\begin{APACrefauthors}%
Scharenborg, O.%
, Norris, D.%
, ten Bosch, L.%
\BCBL {}\ \BBA {} McQueen, J.%
\end{APACrefauthors}%
\unskip\
\newblock
\APACrefYearMonthDay{2005}{}{}.
\newblock
{\BBOQ}\APACrefatitle {How should a speech recognizer work?} {How should a
  speech recognizer work?}{\BBCQ}
\newblock
\APACjournalVolNumPages{Cognitive Science}{29}{}{867--918}.
\PrintBackRefs{\CurrentBib}

\bibitem [\protect \citeauthoryear {%
Schatz%
, Bach%
\BCBL {}\ \BBA {} Dupoux%
}{%
Schatz%
\ \protect \BOthers {.}}{%
{\protect \APACyear {2017}}%
}]{%
schatz2017asr}
\APACinsertmetastar {%
schatz2017asr}%
\begin{APACrefauthors}%
Schatz, T.%
, Bach, F.%
\BCBL {}\ \BBA {} Dupoux, E.%
\end{APACrefauthors}%
\unskip\
\newblock
\APACrefYearMonthDay{2017}{}{}.
\newblock
{\BBOQ}\APACrefatitle {{ASR systems as models of phonetic category perception
  in adults}} {{ASR systems as models of phonetic category perception in
  adults}}.{\BBCQ}
\newblock
\BIn{} \APACrefbtitle {{Proceedings of the 39th Annual CogSci Meeting}.}
  {{Proceedings of the 39th Annual CogSci Meeting}.}
\PrintBackRefs{\CurrentBib}

\bibitem [\protect \citeauthoryear {%
Schatz%
\ \BBA {} Feldman%
}{%
Schatz%
\ \BBA {} Feldman%
}{%
{\protect \APACyear {2018}}%
}]{%
thomashmm}
\APACinsertmetastar {%
thomashmm}%
\begin{APACrefauthors}%
Schatz, T.%
\BCBT {}\ \BBA {} Feldman, N.%
\end{APACrefauthors}%
\unskip\
\newblock
\APACrefYearMonthDay{2018}{}{}.
\newblock
{\BBOQ}\APACrefatitle {{Neural network vs. HMM speech recognition systems as
  models of human cross-linguistic phonetic perception}} {{Neural network vs.
  HMM speech recognition systems as models of human cross-linguistic phonetic
  perception}}.{\BBCQ}
\newblock
\BIn{} \APACrefbtitle {Proceedings of the Conference on Cognitive Computational
  Neuroscience} {Proceedings of the conference on cognitive computational
  neuroscience}\ (\BPGS\ 1--4).
\PrintBackRefs{\CurrentBib}

\bibitem [\protect \citeauthoryear {%
Schatz%
, Feldman%
, Goldwater%
, Cao%
\BCBL {}\ \BBA {} Dupoux%
}{%
Schatz%
\ \protect \BOthers {.}}{%
{\protect \APACyear {To appear}}%
}]{%
schatz2019early}
\APACinsertmetastar {%
schatz2019early}%
\begin{APACrefauthors}%
Schatz, T.%
, Feldman, N.%
, Goldwater, S.%
, Cao, X\BPBI N.%
\BCBL {}\ \BBA {} Dupoux, E.%
\end{APACrefauthors}%
\unskip\
\newblock
\APACrefYearMonthDay{To appear}{}{}.
\newblock
{\BBOQ}\APACrefatitle {{Early phonetic learning without phonetic categories:
  Insights from machine learning}} {{Early phonetic learning without phonetic
  categories: Insights from machine learning}}.{\BBCQ}
\newblock
\APACjournalVolNumPages{Proceedings of the National Academy of Sciences}{}{}{}.
\PrintBackRefs{\CurrentBib}

\bibitem [\protect \citeauthoryear {%
Tiede%
\ \protect \BOthers {.}}{%
Tiede%
\ \protect \BOthers {.}}{%
{\protect \APACyear {2017}}%
}]{%
Haskins}
\APACinsertmetastar {%
Haskins}%
\begin{APACrefauthors}%
Tiede, M.%
, Espy-Wilson, C\BPBI Y.%
, Goldenberg, D.%
, Mitra, V.%
, Nam, H.%
\BCBL {}\ \BBA {} Sivaraman, G.%
\end{APACrefauthors}%
\unskip\
\newblock
\APACrefYearMonthDay{2017}{}{}.
\newblock
{\BBOQ}\APACrefatitle {Quantifying kinematic aspects of reduction in a
  contrasting rate production task} {Quantifying kinematic aspects of reduction
  in a contrasting rate production task}.{\BBCQ}
\newblock
\APACjournalVolNumPages{The Journal of the Acoustical Society of
  America}{141}{5}{3580-3580}.
\newblock
\begin{APACrefURL} \url{https://doi.org/10.1121/1.4987629} \end{APACrefURL}
\newblock
\begin{APACrefDOI} \doi{10.1121/1.4987629} \end{APACrefDOI}
\PrintBackRefs{\CurrentBib}

\bibitem [\protect \citeauthoryear {%
Warstadt%
\ \protect \BOthers {.}}{%
Warstadt%
\ \protect \BOthers {.}}{%
{\protect \APACyear {2019}}%
}]{%
blimp}
\APACinsertmetastar {%
blimp}%
\begin{APACrefauthors}%
Warstadt, A.%
, Parrish, A.%
, Liu, H.%
, Mohananey, A.%
, Peng, W.%
, Wang, S\BHBI F.%
\BCBL {}\ \BBA {} Bowman, S.%
\end{APACrefauthors}%
\unskip\
\newblock
\APACrefYearMonthDay{2019}{}{}.
\newblock
{\BBOQ}\APACrefatitle {BLiMP: A Benchmark of Linguistic Minimal Pairs for
  English} {Blimp: A benchmark of linguistic minimal pairs for english}.{\BBCQ}
\newblock
\APACjournalVolNumPages{arXiv preprint arXiv:1912.00582}{}{}{}.
\PrintBackRefs{\CurrentBib}

\end{thebibliography}

\end{document}